\pdfoutput=1

\documentclass[11pt]{article}

\usepackage{acl}

\usepackage{times}
\usepackage{latexsym}

\usepackage[T1]{fontenc}

\usepackage[utf8]{inputenc}

\usepackage{microtype}

\usepackage{amsmath}
\usepackage{graphicx}
\usepackage{booktabs}
\usepackage{arydshln}
\usepackage{tabularx}
\usepackage{amsfonts}
\usepackage{multirow}
\usepackage{multicol}
\usepackage{cleveref}
\usepackage{xspace}
\usepackage{makecell}
\usepackage{fontawesome5}
\usepackage{colortbl}
\usepackage[ruled, lined, linesnumbered, commentsnumbered]{algorithm2e}
\usepackage{todonotes}
\usepackage{tablefootnote}

\crefformat{section}{\S#2#1#3} 
\crefformat{subsection}{\S#2#1#3}
\crefformat{subsubsection}{\S#2#1#3}
\newcommand{\refequ}[1]{Equation~(\ref{#1})}
\newcommand{\reffig}[1]{Figure~\ref{#1}}
\newcommand{\reftab}[1]{Table~\ref{#1}}

\def\eg{\textit{e.g.}\xspace}

\def\etc{\textit{etc.}\xspace}
\def\ie{\textit{i.e.}\xspace}

\definecolor{myyellow}{RGB}{249, 231, 173}
\definecolor{mygreen}{RGB}{233, 242, 230}

\newcommand{\RNum}[1]{\uppercase\expandafter{\romannumeral #1\relax}}

\makeatletter
\newcommand{\printfnsymbol}[1]{%
  \textsuperscript{\@fnsymbol{#1}}%
}
\makeatother

\title{ACCENT: An \underline{A}utomati\underline{c} Event \underline{C}ommonsense \underline{E}valuatio\underline{n} Me\underline{t}ric for Open-Domain Dialogue Systems}

\author{
Sarik Ghazarian\textsuperscript{\rm 1}\thanks{\indent Equal contribution}\hspace{0.5em}
Yijia Shao\textsuperscript{\rm 2}\printfnsymbol{1}\thanks{\indent The work was done while the author was conducting a summer internship at UCLA.}\hspace{0.5em}
Rujun Han\textsuperscript{\rm 3}\thanks{\indent The collaboration started when the author was a graduate student at USC.}\hspace{0.5em}
\textbf{Aram Galstyan}\textsuperscript{\rm 1}\hspace{0.5em}
\textbf{Nanyun Peng}\textsuperscript{\rm 4} \\
\textsuperscript{\rm 1}University of Southern California / Information Sciences Institute \\
\textsuperscript{\rm 2}Peking University \\
\textsuperscript{\rm 3}AWS AI Labs \\
\textsuperscript{\rm 4}Computer Science Department of University of California, Los Angeles \\
\{sarik, galstyan\}@isi.edu,
shaoyj@pku.edu.cn,
rujunh@amazon.com,
violetpeng@cs.ucla.edu
}

\begin{document}
\maketitle
\begin{abstract}


Commonsense reasoning is omnipresent in human communications and thus is an important feature for open-domain dialogue systems. 
However, evaluating commonsense in dialogue systems is still an open challenge. We take the first step by focusing on {\em event commonsense} that considers events and their relations, and is crucial in both dialogues and general commonsense reasoning. 
We propose \textbf{ACCENT}, an event commonsense evaluation metric empowered by commonsense knowledge bases (CSKBs). ACCENT first extracts event-relation tuples from a dialogue, and then evaluates the response by scoring the tuples in terms of their compatibility with the CSKB. 
To evaluate ACCENT, we construct the first public event commonsense evaluation dataset for open-domain dialogues.
Our experiments show that ACCENT is an efficient metric for event commonsense evaluation, which achieves higher correlations with human judgments than existing baselines.

\end{abstract}

\section{Introduction}
\label{sec:intro}
\begin{figure}[t]
    \resizebox{\columnwidth}{!}{%
    \centering 
    \includegraphics{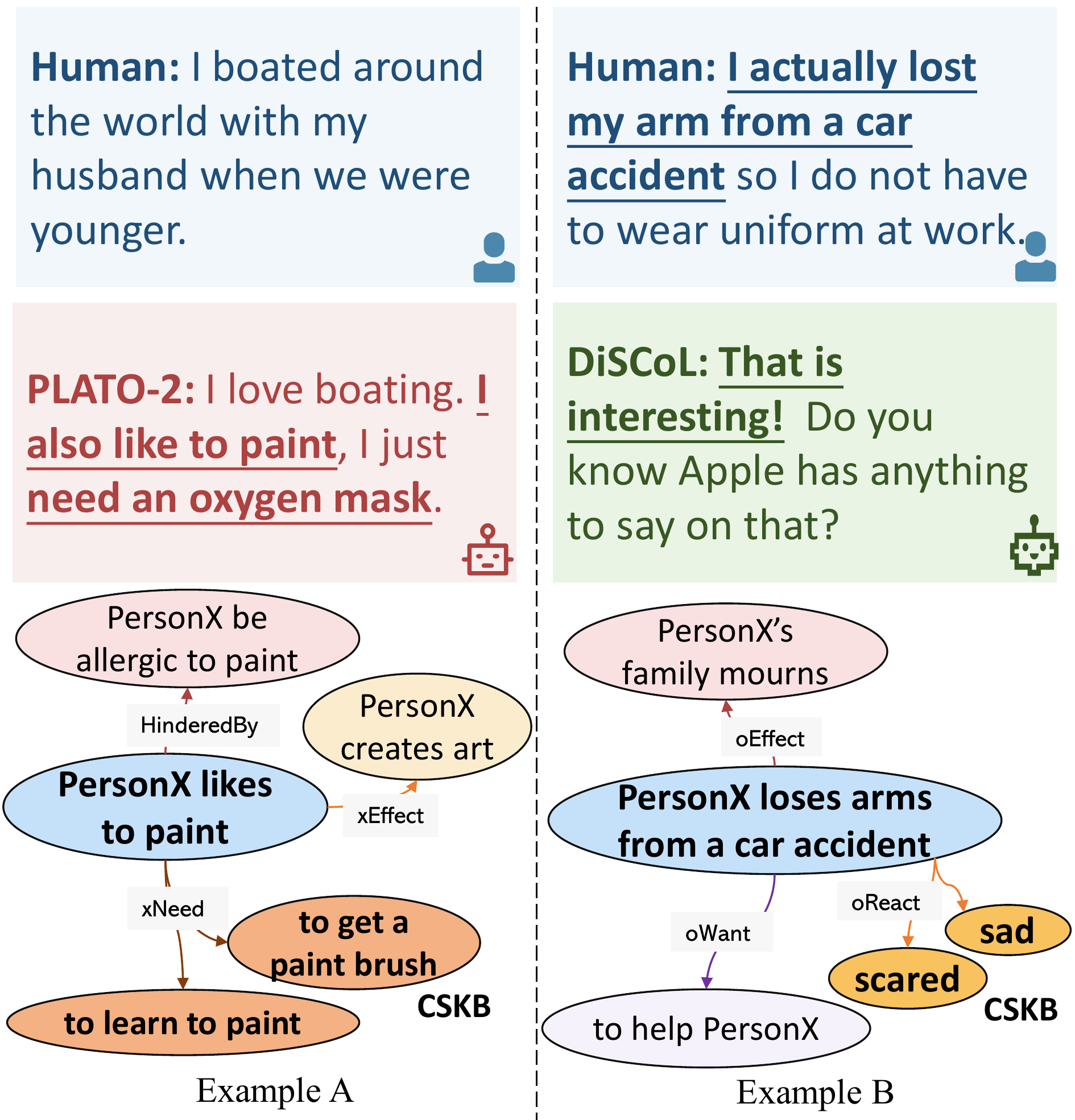}
    }
    \caption{Examples of nonsensical system responses in open-domain dialogues.
    }
    \label{Fig.example}
\end{figure}

Open-domain dialogue systems aim to have natural and engaging conversations with users~\citep{chen2017survey}. The abundance of dialogue corpus~\citep{dziri2018augmenting} and the development of neural models~\citep{radford2019language,lewis-etal-2020-bart} enable  open-domain dialogue systems to generate grammatically correct and meaningful responses 
~\citep{zhang-etal-2020-dialogpt,bao-etal-2021-plato,ghazarian-etal-2021-discol}. 
Despite the success, systems still struggle to consistently produce commonsense-compliant responses as humans do. 
As shown in~\reffig{Fig.example} Example A, the generated response is not compliant with commonsense since ``need an oxygen mask'' is not a reasonable prerequisite for ``like to paint''. Commonsense issues for dialogue systems can also be manifested when we consider the dialogue history.  
For instance, in~\reffig{Fig.example} Example B, the system's response ``\textit{That is interesting!}'' after the user talks about their car accident violates commonly accepted social norms~\citep{frischmann2021common}.

In this work, we study automatic dialogue commonsense evaluation by focusing on \textbf{event commonsense}~\citep{sap-etal-2020-commonsense,rashkin-etal-2018-event2mind}, which concerns commonsense knowledge about events and their relations. Our focus on event commonsense is motivated by the following three observations: 
First, advanced open-domain dialogue systems have been pre-trained on large corpus and thus suffer less from factoid commonsense issues~\citep{petroni-etal-2019-language}. 
Second, events and their relations are key components of commonsense reasoning~\citep{mccarthy1981some}, and our study shows overall commonsense and event commonsense are highly correlated (see \cref{sec:dataset}). Third,  event commonsense aligns well with the interactive nature of open-domain dialogue systems~\citep{huang2020challenges} to complete certain social goals.

To automatically evaluate event commonsense in open-domain dialogues, we propose \textbf{ACCENT}, a reference-free \textbf{A}utomati\textbf{C} Event \textbf{C}ommonsene \textbf{E}valuatio\textbf{N} me\textbf{T}ric which leverages commonsense knowledge bases (CSKBs) and measures the quality of generated responses without having ground-truth reference responses. 
For example, comparing the examples in~\reffig{Fig.example} against the CSKB 
easily reveals commonsense errors in the responses because when ``PersonX likes to paint'', what he/she needs may be ``to get a paint brush'' instead of ``to get an oxygen mask'', and when ``PersonX loses arms from a car accident'', the other person is expected to feel ``sad''. 

While these judgments are intuitive to human, two challenges exist in automating the evaluation process. First, there is a considerable gap between free-form conversational data and the compact commonsense knowledge in the CSKB. 
Second, locating relevant knowledge in the CSKB is non-trivial. 


ACCENT addresses these challenges through a pipeline method that uses an intermediate \textbf{symbolic representation} for commonsense reasoning. 
ACCENT first extracts event-relation tuples from the target response and its preceding dialogue history via a prompt-based generative model trained in a low-resource setting. Those extracted tuples bridge the gap between the free-form dialogue and the compact form of CSKB. Then, a is computed to decide how well each extracted tuple aligns with the CSKB. 

To train and evaluate ACCENT, we construct the first publicly available event commonsense evaluation dataset for open-domain dialogues (see~\cref{sec:dataset}). Besides collecting human commonsense judgments, we request annotators to manually extract event-relation tuples for further analysis. 

Our main contributions are three-fold:
\begin{itemize}
\item{We propose ACCENT, an event commonsense evaluation metric for open-domain dialogue systems. To the best of our knowledge, this is the first work that systematically studies event commonsense in dialogue systems.} %
\item{We construct the first publicly available event commonsense evaluation dataset for open-domain dialogues.\footnote{We release ACCENT and our collected datasets at \url{https://github.com/PlusLabNLP/ACCENT}.} 
}
\item{Extensive experiments show that ACCENT achieves better correlation with human judgments for dialogue commonsense evaluation than several well-designed baselines, and enables easier interpretability of results.} 

\end{itemize}

\section{Background: Event Commonsense}
\label{sec:background}
\begin{figure*}[t]
    \centering 
    \resizebox{0.87\textwidth}{!}{%
    \includegraphics{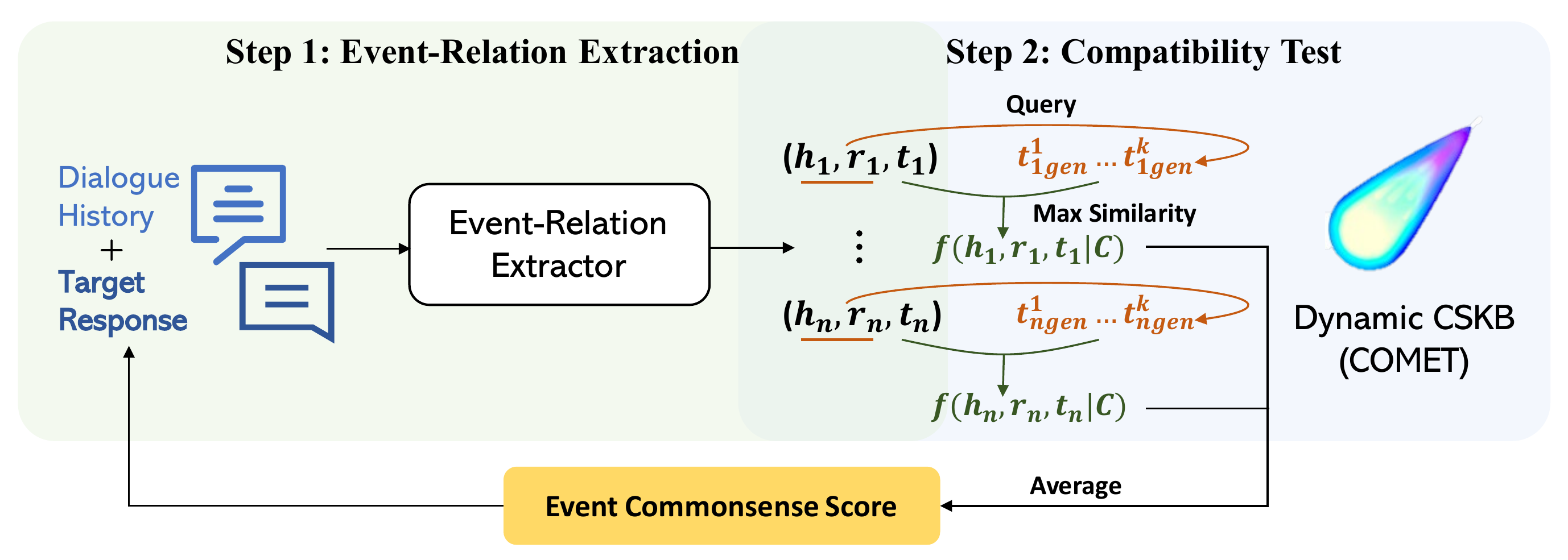}
    }
    \caption{
    The overview of ACCENT. Given the target response and its dialogue history, ACCENT first extracts the event-relation tuples. Then, the compatibility test (detailed illustration in~\reffig{Fig.compatibility}) assigns a score to each tuple: ACCENT queries the dynamic CSKB, \ie, COMET, with $h$ and $r$, and generates $k$ events. The compatible score is the maximum similarity between the ground-truth $t$ and the $k$ generated events $\{t_{gen}^i\}_{i=1}^{k}$. Scores for all tuples in a response are averaged to obtain the event commonsense score for the target response.}
    
    \label{Fig.method}
\vspace{-0.5em}
\end{figure*}

\begin{figure}[!h]
    \resizebox{\columnwidth}{!}{%
    \centering 
    \includegraphics{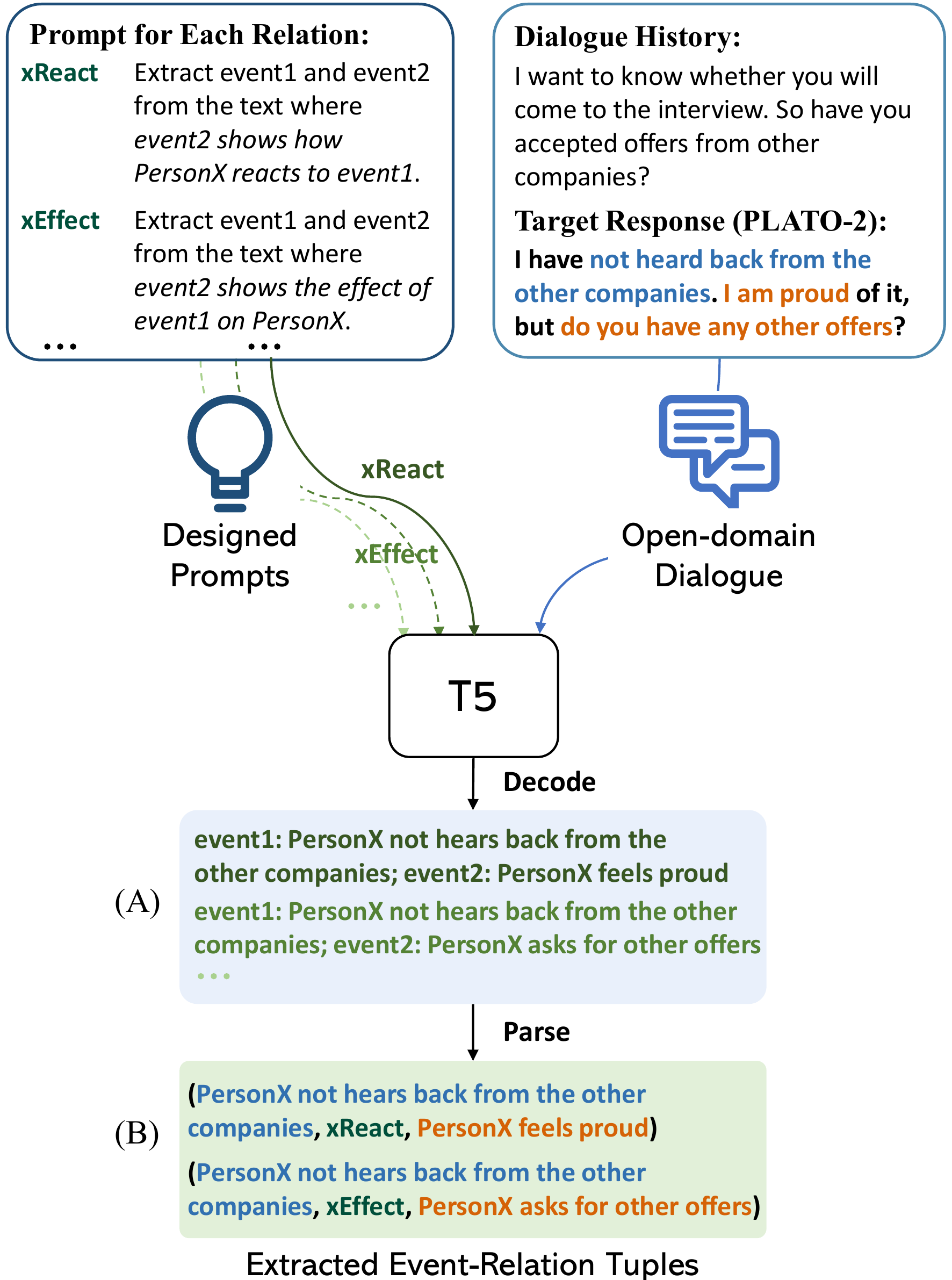}
    }
    \caption{Illustration for event-relation extraction. For each relation $r\in\mathcal{\tilde{R}}$, we use its corresponding prompt to guide the model to generate $h$ and $t$. 
    The final tuple is parsed from the generated output.}
    \label{Fig.eventualization}
\vspace{-0.5em}
\end{figure}

Endowing machines with human-like commonsense reasoning capabilities has been an ultimate goal of artificial intelligence research for decades~\citep{mccarthy1981some,lecun2022path}. 
While many early works focused on factoid commonsense or the knowledge about concepts~\citep{lenat1995cyc,liu2004conceptnet}, event commonsense emerges as an important aspect for machine commonsense measurement~\citep{chen-etal-2021-event}. Compared with concepts or entities, events are more informative, involving actions, participants, time \etc Besides, event commonsense also requires understanding various relations between events~\citep{kuipers1984commonsense,rashkin-etal-2018-event2mind} which would facilitate complex reasoning, especially in interactive scenarios such as dialogues.

Among the current commonsense resources (related works in Appendix~\ref{appendix:cskb}), $\text{ATOMIC}_{20}^{20}$~\citep{Hwang2021COMETATOMIC2O} is a comprehensive CSKB including physical-entity, event-centered, and social-interaction knowledge. Its event-centered and social-interaction components take up 84.4\% tuples of the entire knowledge base, providing knowledge regarding how events/human actions are associated with other events/actions. For example, given the event ``X runs out of stream'', according to $\text{ATOMIC}_{20}^{20}$, this event may happen after ``X exercises in the gym'', and the person X is likely to ``feel tired''. 

\section{Method}
\label{sec:method}

We present ACCENT, as a framework for event commonsense evaluation. \reffig{Fig.method} gives an overview of ACCENT with two major components. 

\subsection{Symbolic Intermediate Representation}
\label{sec:preliminaries}

ACCENT uses event-relation tuples as the symbolic intermediate representation. Each tuple contains a head event and a tail event which are connected through an event relation. We formally define events and relations below.

\noindent
\textbf{Event}\quad
Following~\citet{pustejovsky2003timebank}, we define events as short phrases with a trigger word and its arguments (\eg, I like to paint). 
To better align with $\text{ATOMIC}_{20}^{20}$, we normalize the event by replacing tokens referring to people with \texttt{Person} variable (\eg, PersonX likes to paint).

\noindent
\textbf{Relation}\quad
We select $\mathcal{\tilde{R}}=$\{\text{xIntent}, \text{xWant}, \text{oWant}, \text{xReact}, \text{oReact}, \text{xNeed}, \text{xAttr}, \text{xEffect}, \text{oEffect}, \text{HinderedBy}, \text{IsAfter}, \text{HasSubEvent}\}\footnote{``x'' and ``o'' pertain to PersonX and other person(s).} from $\text{ATOMIC}_{20}^{20}$ relations. These relations cover human behaviors, \ie, motivation, want, reaction, need, description, towards events~\citep{sap-etal-2019-social}, the cause-effect and constraint in force dynamic~\citep{talmy1988force}, the temporal information, as well as the parent-child relation in event hierarchy. Examples for each relation are in Appendix~\ref{appendix:relation}.



\subsection{Event-Relation Extraction}
\label{sec:eventualization}

The input of the event commonsense evaluation task is a list of utterances $\{u_0, u_1, \cdots, u_{n-1}\}$ representing the dialogue history and the target response $u_n$. ACCENT first converts the free-form text into event-relation tuples. To retain the information in $u_n$, ACCENT extracts tuples whose head and tail events are both from the target response (denoted as ``Single''). Besides, to capture event commonsense issues conditioned on the dialogue history (\eg, \reffig{Fig.example} Example B), ACCENT also extracts tuples whose two events come from $u_n$ and $u_{n-1}$ respectively (denoted as ``Pair'').

As illustrated in~\reffig{Fig.eventualization}, the event-relation extractor in ACCENT is a T5 model $\mathcal{M}$~\citep{2020t5} guided to generate the head and tail events via designed prompts for each relation. ACCENT concatenates the prompt for $r\in\mathcal{\tilde{R}}$ and the dialogue as the input and fine-tunes $\mathcal{M}$ in a low resource setting. When the relation $r$ exists in the input utterances, the fine-tuned $\mathcal{M}$ is expected to generate the head and tail events following a particular format, \ie, ``event1: \{head\}; event2: \{tail\}'', so that the tuple can be parsed from the decoded sequence (from Block A to Block B in~\reffig{Fig.eventualization}). Otherwise, the fine-tuned $\mathcal{M}$ is expected to output ``None''. For each $r\in\mathcal{\tilde{R}}$, the designed prompt explains the semantic meaning of $r$ and triggers the model to generate the head and tail events (the prompts are included in Appendix~\ref{appendix:relation}). 
At the inference time, we query $\mathcal{M}$ with prompts for each $r$ and parse the generated outputs to get $h$ and $t$ to construct tuples.


\label{sec:compatibility}
\begin{figure}[t]
    \resizebox{\columnwidth}{!}{%
    \centering 
    \includegraphics{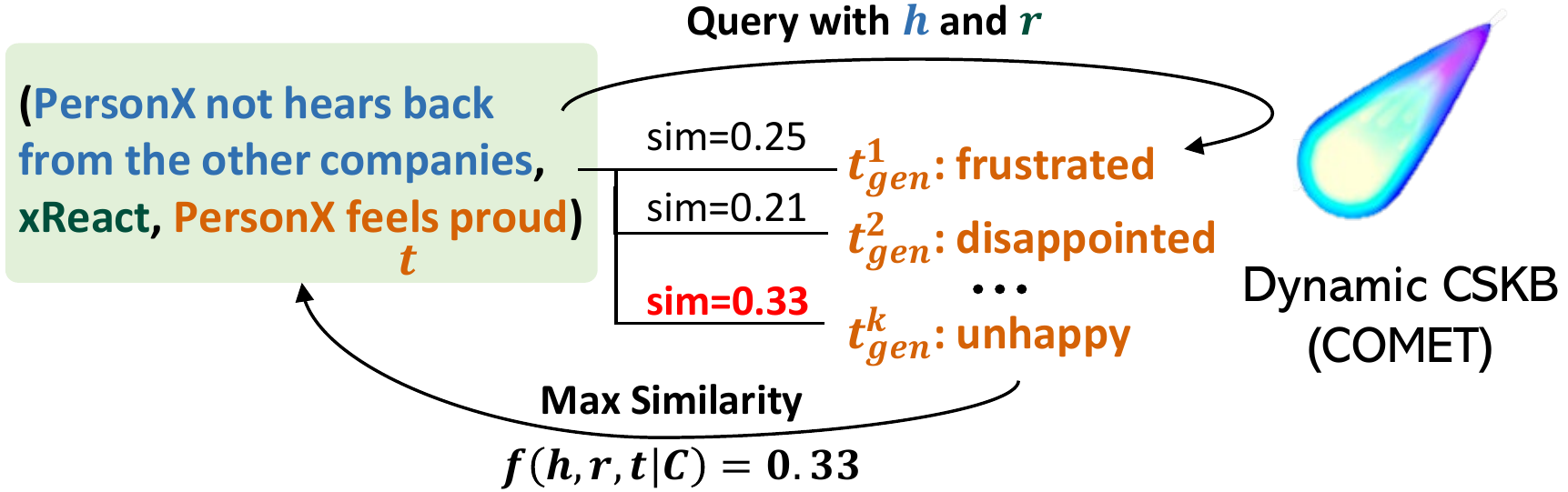}
    }
    \caption{An example of compatibility test. We query the dynamic CSKB with $h$ and $r$, and the compatibility score is the maximum similarity between $t$ and the generated tail events ($\{t_{gen}^i\}_{i=1}^{k}$).}
    \label{Fig.compatibility}
\vspace{-1em}
\end{figure}

\subsection{Compatibility Test}

After extracting event-relation tuples, ACCENT checks whether these tuples are sensible through a compatibility test. Denoting the CSKB as $\mathcal{C}$, 
the compatibility test aims to learn a scoring function $f$ based on $\mathcal{C}$, where $f((h,r,t)|\mathcal{C})$ represents the compatibility of the target tuple $(h, r, t)$ with the CSKB $\mathcal{C}$. we propose to score $(h, r, t)$ by querying a \textit{dynamic version of $\mathcal{C}$} with $h$ and $r$. ~\reffig{Fig.compatibility} gives an example of the whole process.




Specifically, ACCENT uses COMET~\citep{bosselut-etal-2019-comet} as the dynamic CSKB. COMET adapts the pre-trained language model by fine-tuning it on $\mathcal{C}$ through a conditional generation task where ``\{head\} \{relation\} [GEN]'' is the source and a tail event is the target. To score $(h, r, t)$, we query the model by requiring it to generate $t_{gen}$ given ``\{$h$\} \{$r$\} [GEN]''. The beam search method is applied for decoding, so we obtain a set of generated tail events, $\{t_{gen}^i\}_{i=1}^{k}$, where $k$ is the beam size.

The compatibility score for $(h, r, t)$ is then computed by checking the similarity between $t$ and the most similar $t_{gen}$ among $\{t_{gen}^i\}_{i=1}^{k}$:

\begin{equation}
\small
\begin{aligned}
    f((h, r, t)|\mathcal{C}) \! = \! \max_{1\leq i \leq k} \! \cos(\operatorname{embed}(t), \operatorname{embed}(t_{gen}^i))
\label{eq:compatibility}
\end{aligned}
\end{equation}
Here, $\operatorname{embed}(\cdot)$ is parameterized by a Sentence-BERT model~\citep{reimers-gurevych-2019-sentence}. 

After getting the compatibility scores for each extracted tuple, we average them to get the final score for the target response (see~\reffig{Fig.method}).

\section{Data Collection}
\label{sec:dataset}
We construct the first event commonsense evaluation dataset for open-domain dialogues through crowdsourcing on Amazon Mechanical Turk (MTurk). In this section, we describe the collection procedure and the details of the dataset.


\subsection{Dialogue Data Preparation}
\label{sec:data_preparation}
We select dialogue histories from DailyDialog~\citep{li-etal-2017-dailydialog}, PersonaChat~\citep{zhang-etal-2018-personalizing}, and TopicalChat~\citep{Gopalakrishnan2019} \textit{human-human} dialogues. 
The dialogue history is limited to at most 4 consecutive utterances. 
Since human utterances barely contradict event commonsense, to better evaluate machine generated dialogues, we collect responses using 
advanced dialogue systems, DialoGPT~\citep{zhang-etal-2020-dialogpt}, PLATO-2~\citep{bao-etal-2021-plato}, DiSCoL~\citep{ghazarian-etal-2021-discol}.

To ensure most samples contain events and are meaningful for event commonsense evaluation, we filter samples using the following criteria: (1) the response contains at least 5 words; (2) the response contains at least 1 non-interrogative sentence\footnote{We check this by finding sentences that are not ended with a question mark (``?'').}; (3) the response is more than a courtesy (\eg, ``It's been nice chatting with you.'')\footnote{These responses are manually filtered out.}. 
After filtering, we randomly select 300 samples and split them into 200 for training and 100 for testing. We name this dataset \textbf{DECO} (\textbf{D}ialogue \textbf{E}vent \textbf{Co}mmonsense Dataset).

\subsection{Tuple Extraction}
To train the event-relation extractor of ACCENT, we collect human extracted event-relation tuples from DECO training set. Annotators are shown with the target response, the dialogue history, a specific relation, and are requested to 
compose event-relation tuples. They are allowed to tick ``I cannot find any tuple'' if no tuple can be found. We also request them to select whether the tuple belongs to ``Single'' or ``Pair'' (defined in~\cref{sec:eventualization}) for each tuple they extract. \reffig{Fig.amt} in Appendix~\ref{appendix:data} shows our data collection panel. We launched HITs\footnote{HIT is an assignment unit on Amazon MTurk.} for relations in $\mathcal{\tilde{R}}$ repeatedly until we obtained at least 20 tuples for each relation. 
In order to ensure the test set is comprehensive, we particularly request annotators to compose tuples for all 12 relations in $\mathcal{\tilde{R}}$ (100 samples $\times$ 12 relations in total).

\label{sec:tuple_collection}
\begin{figure}[t]
    \resizebox{\columnwidth}{!}{%
    \centering 
    \includegraphics{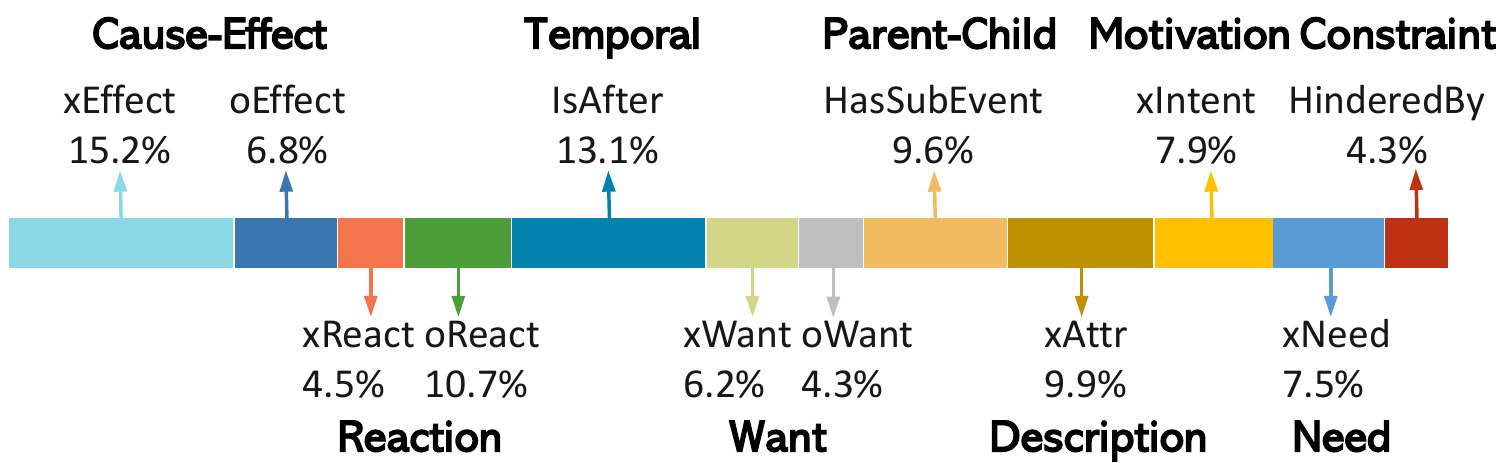}
    }
    \caption{Relation distribution in DECO test set.}
    \label{Fig.relation}
\vspace{-1em}
\end{figure}

A separate validation round was conducted to check whether each extracted tuple satisfies (1) the head and tail are events, (2) the head and tail come from $u_n$ or $u_{n-1}$, (3) the relation between the head and tail can be inferred from the dialogue. 
A tuple is deemed valid if the majority of 3 annotators vote ``yes''. 
After removing invalid tuples (the dialogue numbers remain unchanged), we collected 307 tuples for training and 467 tuples from the DECO test set. 
\reffig{Fig.relation} shows the relation distribution in the densely annotated test set. More details about DECO statistics are included in Appendix~\ref{appendix:data}.

\subsection{Commonsense Scoring}
\label{sec:scoring_collection}
We instruct annotators to score target responses in terms of event commonsense by focusing on the events and their relations (the guideline is shown in~\reffig{Fig.scoring_guideline}). 
Each response was annotated by 3 individual annotators with a scale of 1 to 5. 
Following~\citet{mehri-eskenazi-2020-unsupervised}, we measure the inter annotator agreement (IAA) by correlating each annotation with the mean of the other annotations for the same sample, and the Spearman correlation is 0.578 showing an acceptable agreement.\footnote{0.40-0.69 implies strong relationship.} 
The final event commonsense score assigned to each sample is the average of 3 individual ratings. \looseness=-1

We also requested the annotators to judge the overall commonsense of a dialogue response
before introducing event commonsense to annotators. Among the 900 annotation pairs we collected, the Spearman correlation between the two scores reaches 0.862, 
which indicates that \textit{event commonsense is a key component in overall commonsense reasoning.} \looseness=-1 

\subsection{Additional Human-Machine Dialogues}
We further explore the generalization ability of ACCENT on responses with \textit{human-machine} dialogue histories.
We select 100 samples from ConTurE~\citep{ghazarian-etal-2022-wrong}, a turn-level evaluation dataset, to annotate event commonsense scores. We denote this dataset as \textbf{ConTurE Subset}. Its statistics are also included in Appendix~\ref{appendix:data}.

\section{Experiments}
\label{sec:experiment}

\subsection{Setups}
\label{sec:setup}
We compare ACCENT with baseline methods for event commonsense evaluation and also examine its two components separately. Therefore, our experiments include three setups for the evaluation: 

\noindent
\textbf{Setup 1 (Metrics Performance)}\quad
Our main goal is to evaluate the commonsense metric, and we achieve this by computing the correlation between automatic scores and human judgments. 
ACCENT and baseline metrics are tested on DECO test set and ConTurE Subset. 

\noindent
\textbf{Setup 2 (Event-Relation Extraction)}\quad
We evaluate the performance of the event-relation extraction component of ACCENT by comparing the automatically extracted tuples with human extracted tuples on DECO test set. 
We view checking whether a tuple with relation $r$ is extracted from the utterances $u_n$ and $u_{n-1}$ as a binary classification problem and compute the F1 score. We also measure how ``close'' the automatically extracted head and tail events are to human extraction results. We convert the tuple into a sentence by concatenating the head and tail, and then compute BLEU-2~\citep{papineni-etal-2002-bleu} and BERTScore~\citep{Zhang*2020BERTScore:}.

\noindent
\textbf{Setup 3 (Compatibility Test):}\quad
The compatibility test component of ACCENT can be viewed as a tuple scoring task. We compare our proposed approach with other tuple scoring methods on a 
large-scale benchmark~\citep{fang-etal-2021-benchmarking} which contains event-relation tuples with 0 (compatible to a given CSKB) or 1 (not compatible to the CSKB) scores.
Since the training relations in this benchmark differ from relations supported by the off-the-shelf COMET, we train our own COMET on its training set (see Appendix~\ref{sec:population_detail} for more details) to make our compatibility test component applicable to this test set. This benchmark dataset covers all 12 relations in $\mathcal{\tilde{R}}$ as well as 6 more relations.  

\subsection{Baselines}
\label{sec:baselines}
We compare ACCENT with 5 baseline metrics: 
(1, 2)  \textbf{FED-understandable}/\textbf{appropriate}~\citep{mehri-eskenazi-2020-unsupervised} are two off-the-shelf baselines.  ``Understandable'' and ``Semantically Appropriate'' are closer to commonsense compared to the rest of the criteria in FED. (3) \textbf{Cross-encoder} is a widely used model for sentence-pair regression tasks. We use BART~\citep{lewis-etal-2020-bart} as the backbone. (4) \textbf{Cross-encoder (COMET)} is a variant of (3) with COMET trained on $\text{ATOMIC}_{20}^{20}$ as the backbone. 
(5) \textbf{MLP regressor}~\citep{zhou-etal-2021-commonsense} is trained with neural features from DialoGPT and symbolic features from ConceptNet (details in~\cref{sec:related_work}). 
The cross-encoders and the MLP regressor require event commonsense scores to train the model in an end-to-end manner. We use the annotated scores in DECO training set to train them, and split 20\% data for validation to conduct hyperparameter search.

For Setup 2, we consider the following baseline approaches: (1) \textbf{ASER Extractor}~\citep{ZhangLPSL20} first extracts events through patterns from dependency parsing and then uses a neural classifier to predict the relation. (2) \textbf{CSKB Search}~\citep{zhou-etal-2021-commonsense} searches the one-hop neighbors in $\text{ATOMIC}_{20}^{20}$ through keyword matching.  

For Setup 3, we consider 4 tuple scoring baselines. These baselines convert a tuple to an embedding and train a binary classifier to give score: (1) \textbf{\textsc{Bert}} feeds $h, r, t$ to \textsc{Bert} and concatenates their [CLS] embeddings to get the tuple embedding. (2)\textbf{\textsc{BertSAGE}}~\citep{fang2021discos} further concatenates the average embedding of the neighbors of $h$ and $t$ in an event knowledge graph. (3) \textbf{\textsc{KG-Bert}}~\citep{yao2019kg} inputs ``[CLS], $h$, [SEP], $r$, [SEP], $t$'' to get the tuple embedding. (4) \textbf{\textsc{KG-BertSAGE}}~\citep{fang-etal-2021-benchmarking} further concatenates the average embedding of neighboring nodes. 
We use $\text{RoBERTa}_\text{LARGE}$~\citep{liu2020roberta} as the backbone which has roughly the same parameter budget with COMET to have a fair comparison. 

The details of the baseline implementations are in Appendix~\ref{appendix:baseline}.

\subsection{ACCENT Implementation}
\label{sec:implementation}
The proposed ACCENT framework is implemented using the Transformers library~\citep{wolf-etal-2020-transformers}. For event-relation extraction, we fine-tune T5-base\footnote{\url{huggingface.co/t5-base}} for 50 epochs with the batch size of 4 and the learning rate of 5e-5. The training data comes from the human extracted tuples from DECO training set. We additionally select 5 negative samples (dialogues that do not have a certain relation) 
per relation from the training set and set their target output as ``None'' to guide the model to handle cases which do not contain a certain relation. During inference, if no tuple is extracted after considering all relations, we assign a score of 0.5 to the sample. For compatibility test, we use the off-the-shelf COMET model trained on $\text{ATOMIC}_{20}^{20}$~\citep{Hwang2021COMETATOMIC2O}\footnote{\url{github.com/allenai/comet-atomic-2020}}. When querying COMET through generation, we use beam search with a beam size of 10 to get commonly sensible tail events. The $\operatorname{embed}(\cdot)$ in \refequ{eq:compatibility} is parameterized by \texttt{paraphrase-MiniLM-L6-v2} provided in the Sentence-Transformers library\footnote{\url{huggingface.co/sentence-transformers/paraphrase-MiniLM-L6-v2}}.

\section{Results and Analysis}
\label{sec:analysis}

\subsection{Metrics Performance}
\label{sec:metric_results}

\begin{table}
\centering
\small
\renewcommand{\tabcolsep}{1.8mm}
\begin{tabular}{lcc|cc} 
\toprule
                               & \multicolumn{2}{c|}{\textbf{DECO}}                                   & \multicolumn{2}{c}{\textbf{ConTurE}}                                     \\
                               & \multicolumn{1}{c}{\textbf{$\gamma$}} & \multicolumn{1}{c|}{\textbf{$\rho$}} & \multicolumn{1}{c}{\textbf{$\gamma$}} & \multicolumn{1}{c}{\textbf{$\rho$}}  \\ 
\midrule
FED-appropriate &   -0.16  & -0.10  &  -0.09 & -0.04    \\
FED-understandable  &  -0.12 & -0.07   & -0.08 & -0.04 \\
Cross-encoder   &   ~0.15   & ~0.15  &    -0.05 & -0.09 \\
Cross-encoder (COMET)   &   ~0.17 & ~0.17 &  ~0.00     &  ~0.00\\
MLP Regressor &  ~0.11  & ~0.01  & ~0.17    & ~0.16        \\
\hline
ACCENT (Ours)             &  ~\textbf{0.30}    & ~\textbf{0.30}             &     ~\textbf{0.21} &   ~\textbf{0.22}                         \\
\bottomrule
\end{tabular}
\caption{Pearson ($\gamma$) and Spearman ($\rho$) correlations between human judgments and different automatic evaluation metrics. The results for ACCENT are all significant ($p < 0.05$). 
}
\label{table:main}
\vspace{-0.5em}
\end{table}

\begin{table*}
\centering
\resizebox{\textwidth}{!}{%
\begin{tabular}{lllll} 
\toprule
\textbf{Dialogue History}                                     & \textbf{Target Response}                                                               & \textbf{Human} & \textbf{ACCENT} & \textbf{Extracted Tuple}                                                  \\ 
\midrule
\makecell[l]{Yes, I run them all the time!\\How exciting, are you ready?}    & \makecell[l]{I'm ready for the marathon! So I'm going to\\be running a half marathon in a few weeks.} & 0.80           & 0.61            & \makecell[l]{(PersonX is ready for the marathon,\\xEffect, PersonX run a half marathon)}  \\
\midrule[0.5pt]
\makecell[l]{I like pizza as long as it's a\\meat lover. I hate vegetables.} & Lol,  yes it is! Great for you to eat so healthy!                                      & 0.53          & 0.45            & \makecell[l]{(PersonX hate vegetables, xAttr,\\PersonX be healthy)}                       \\
\midrule[0.5pt]
I had an accident. & \makecell[l]{That's what I like to call the most beautiful\\thing.} & 0.27 & 0.23 & \makecell[l]{(PersonX has an accident, oWant, PersonY\\wants to call the most beautiful thing)} \\
\bottomrule
\end{tabular}
}
\caption{Examples of evaluation scores given by ACCENT from the DECO test set. Only $u_{n-1}$ in the dialogue history is shown, and the human annotations are normalized to the range of $(0, 1)$ for better demonstration. ``Extracted Tuple'' column shows tuples extracted in ACCENT framework which account for the given scores.}
\label{table:case_study}
\end{table*}

\reftab{table:main} shows the correlations between automatic scores and human annotations. ACCENT uniformly outperforms the baselines on both two test sets. Specifically, off-the-shelf metrics (``FED-appropriate'', ``FED-understandable'') perform poorly. 
For ``Cross-encoder (COMET)'', its results show that implicitly using the CSKB through transfer learning cannot yield satisfactory performance. Besides, cross-encoders 
fail to generalize well to ConTurE Subset whose dialogue histories are from human-machine dialogues. For ``MLP Regressor'', although it tries to utilize the CSKB explicitly, it is not as effective as ACCENT.

Some examples from the DECO test set and their event commonsense scores given by ACCENT are shown in~\reftab{table:case_study}. These scores are close to human judgements and enjoy great interpretability owning to the extracted event-relation tuples.

\begin{figure}[t]
\centering 
    \resizebox{0.9\columnwidth}{!}{%
    \includegraphics{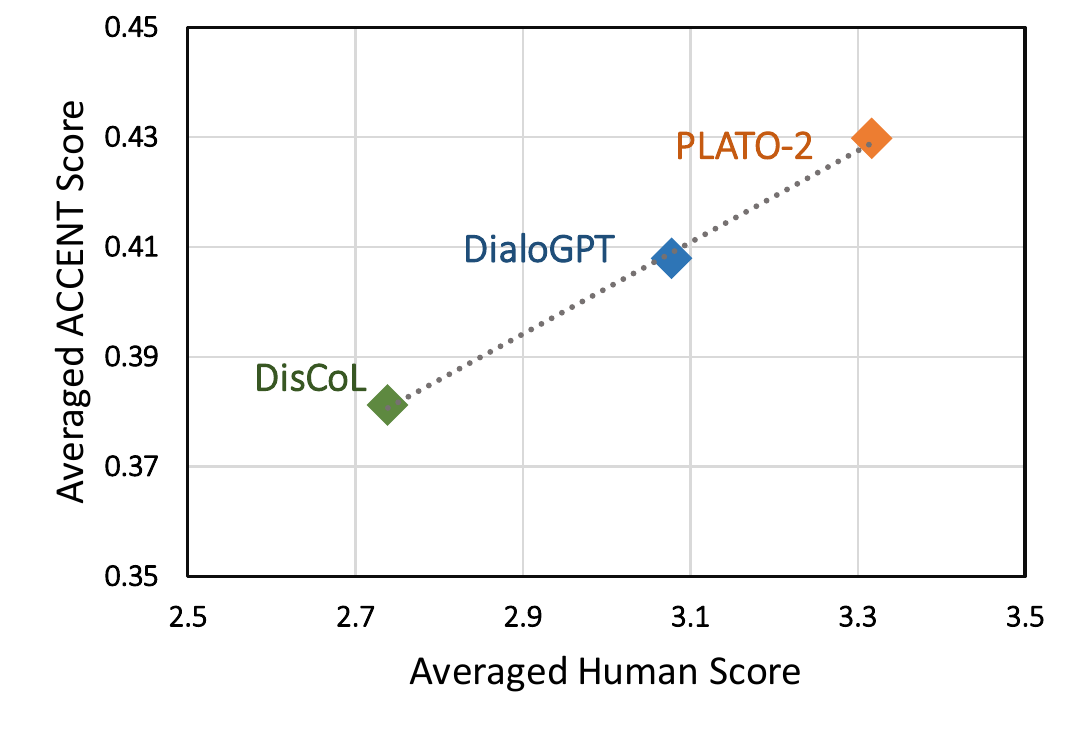}
    }
    \caption{Average event commonsense scores of generated responses of different models using human annotations (scale 1 to 5) and ACCENT automatic evaluation (scale 0 to 1). The rankings of systems given by human and ACCENT are the same.}
    \label{Fig.toy}
\vspace{-0.5em}
\end{figure}

Apart from the sample-level correlation, we further examine whether ACCENT can reflect model performance in terms of event commonsense. \reffig{Fig.toy} shows the rankings of three dialogue systems used in DECO construction given by human and ACCENT. Human and ACCENT rank the three systems exactly the same and the two sets of averaged scores highly correlates with each other.

\subsection{Tuple Extraction Performance}
\label{sec:eventualization_result}
\begin{table}
\centering
\resizebox{\columnwidth}{!}{%
\begin{tabular}{lccccc} 
\toprule
                & \textbf{P} & \textbf{R}    & \textbf{F1} & \textbf{BLEU} & \textbf{BERTScore} \\ 
\midrule
CSKB Search     &  29.9  & \textbf{96.3}   & \textbf{45.7}            &  26.9            &89.8       \\
ASER Extractor  & \textbf{31.5}  & 23.6  &  27.0           & 32.4   & 89.3     \\ 
\midrule
\textbf{Ours}       & 31.4  & 55.0  & 40.0            & \textbf{41.6}       & \textbf{93.5}       \\
\bottomrule
\end{tabular}
}

\caption{Performances of different event-relation extraction methods on DECO test set. P: Precision. R: Recall.}
\label{table:eventualization}
\vspace{-0.5em}
\end{table}

\reftab{table:eventualization} shows the results of Setup 2 where we evaluate the event-relation extraction performance on DECO test set. Our proposed method achieves much higher BLEU and BERTScore than two baselines, indicating that the composed events in tuples have reasonable quality. However, joint event-relation extraction remains challenging because it combines the event extraction and relation identification. Although our proposed method has higher score than ASER Extractor by F1, it still has plenty of room for improvement. 
As for CSKB Search, 
it usually returns a lot of tuples, thus resulting in high recall and very poor precision. Also, searching CSKB is not applicable in our framework because this method can only return sensible tuples.

\subsection{Compatibility Test Performance}
\begin{table}
\centering
\resizebox{0.75\columnwidth}{!}{%
\begin{tabular}{lcc} 
\toprule
              & \textbf{Subset} & \textbf{All}    \\ 
\midrule
\textsc{Bert}          & 62.0\small{$\pm$0.3}           & 61.5\small{$\pm$0.3}           \\
\textsc{BertSAGE}      & 55.8\small{$\pm$0.7}           &       55.8\small{$\pm$0.7}      \\
\textsc{KG-Bert}       & 62.6\small{$\pm$0.7}           & 62.3\small{$\pm$0.8}           \\
\textsc{KG-BertSAGE}   & 63.2\small{$\pm$0.4}           & 62.9\small{$\pm$0.3}           \\
\hline
\textbf{Ours} & \textbf{68.0}\small{$\pm$0.8}  & \textbf{67.6}\small{$\pm$0.8}  \\
\bottomrule
\end{tabular}
}

\caption{Test results on the CSKB compatibility benchmark. We report the overall AUC across all relations (``All'') and the AUC across samples with our target relations (``Subset''). Both the averaged metric and its standard deviation are reported over 3 runs.}
\label{table:population}
\end{table}

\reftab{table:population} depicts the test results on the benchmark dataset. Our method outperforms all baselines, and it does not require negative samples for training. 
The major difference between our method and those tuple scoring baselines is that we use the tuples in the existing CSKB to train a dynamic CSKB, \ie, COMET, instead of a discriminative model. We assume our strong results may be due to the generalization ability of the dynamic CSKB.

\subsection{Ablation Studies}
\label{sec:ablation}

\begin{table}
\centering
\resizebox{0.85\columnwidth}{!}{%
\begin{tabular}{l|lcc} 
\toprule
\multicolumn{1}{l}{} &                      & \textbf{DECO} & \textbf{ConTurE}  \\ 
\hline
\multicolumn{1}{l}{} & \textbf{ACCENT (whole)} &  \textbf{0.30}    & \textbf{0.22}            \\ 
\hline
\multirow{4}{*}{\RNum{1}}   & ~ \small{ASER Extractor}     &     0.14          &    0.00               \\
                     & ~ \small{w/o Pair}           &   0.19            &   0.08                \\
                     & ~ \small{w/o Single}         &   0.24            &   0.18                \\
                     & {\cellcolor[rgb]{0.925,0.925,0.925}}~ \small{Gold Tuples}        & {\cellcolor[rgb]{0.925,0.925,0.925}}\textbf{0.42}              & {\cellcolor[rgb]{0.925,0.925,0.925}} -                   \\ 
\hline
\multirow{3}{*}{\RNum{2}}   & ~ \small{Bert}     & -0.08              &     0.09              \\
                     & ~ \small{KG-Bert}            &  0.13             &   0.19                \\
                     & ~ \small{COMET (neural)}     &  0.16             &   0.05                \\
\bottomrule
\end{tabular}
}
\caption{Ablation results measured by Spearman correlation.  \RNum{1}: Ablation of the event-relation extraction part. The gray row shows the results using human extracted tuples which provides an upper bound. \RNum{2}: Ablation of the compatibility test part of ACCENT.}
\label{table:ablation}
\vspace{-0.5em}
\end{table}

We conduct ablation studies to explore (1) whether the proposed event-relation extraction method can lead to better final metric performance; (2) given the automatically extracted tuples, whether the proposed compatibility test method can lead to higher correlation with human judgment. 

To answer (1), we 
compare different methods to get the event-relation tuples (Part \RNum{1} in~\reftab{table:ablation}). Among the event-relation extraction baselines, we only consider ASER Extractor because CSKB search is not applicable in our framework as discussed in~\cref{sec:eventualization_result}. Note that the event-relation extractor in ACCENT considers tuples 
in both ``Single'' and ``Pair'' settings to cover two potential types of errors (see~\cref{sec:eventualization}). To verify this, we compare the variations of our proposed method where we only use tuples marked as ``Single'' or ``Pair'' for model training. Also, the human extracted tuples in DECO test set are used to provide an upper bound.


To answer (2), we fix the event-relation extraction part and change the compatibility test part (Part \RNum{2} in~\reftab{table:ablation}). We consider \textbf{\textsc{Bert}} and \textbf{\textsc{KG-Bert}} trained on the CSKB compatibility benchmark because they do not need event graph information and can be seamlessly applied to our compatibility test. Also, 
while we query COMET through tail generation, another intuitive design is using the model loss with ``\{h\} \{r\} [GEN]'' as the source and $t$ as the target to give scores. We map the loss to $(0, 1)$ through an exponential function, and name this alternative as ``\textbf{COMET (neural)}'' for it skips the symbolic decoding of $t_{gen}$. 

\reftab{table:ablation} demonstrates that the whole ACCENT gives the best result. 
Considering the variations of our design, ``w/o Pair'' gives much lower results, indicating that limiting the symbolic intermediate representation to only the information contained in the target response is not enough. This observation is in accord with our finding that some event commonsense errors occur when we take the dialogue history into account.

\begin{table}
\centering
\resizebox{0.85\columnwidth}{!}{%
\begin{tabular}{lc|cc} 
\toprule
                  & \textbf{STS Avg.} & \textbf{DECO} & \textbf{ConTurE}  \\ 
\midrule
Sentence-BERT     & 79.82             & 0.30          & 0.22              \\
DiffCSE\tablefootnote{\url{https://huggingface.co/voidism/diffcse-roberta-base-sts}}           & 78.21             & 0.12          & 0.25              \\
ESimCSE\tablefootnote{\url{https://huggingface.co/ffgcc/esimcse-roberta-base}}           & 77.44             & 0.19          & 0.24              \\
Sup-SimCSE\tablefootnote{\url{https://huggingface.co/princeton-nlp/sup-simcse-roberta-base}} & \textbf{82.52}    & \textbf{0.31} & \textbf{0.26}     \\
\bottomrule
\end{tabular}
}
\caption{Results with different sentence embedding methods measured by Spearman correlation. Following~\citet{gao-etal-2021-simcse}, we use the average results on the semantic textual similarity (STS) tasks to reflect the sentence embedding performance.}
\label{table:embedding}
\end{table}

Another empirical discovery is that although ``COMET (neural)'' is a direct way of using the dynamic CSKB, its performance is poorer than what we propose in ACCENT. We assume that comparing $t$ and $t_{gen}$ in a symbolic fashion can yield more comparable scores among tuples with different relations (details in Appendix~\ref{appendix:compatibility}). 

In our implementation of ACCENT, the comparison of $t$ and $t_{gen}$ is done by calculating the cosine similarity between their Sentence-BERT embeddings. We further experiment with other sentence embedding methods based on contrastive learning. 
Specifically, we consider DiffCSE~\citep{chuang-etal-2022-diffcse}, ESimCSE~\citep{wu-etal-2022-esimcse} which are two unsupervised contrastive learning frameworks for learning sentence embeddings. We also consider Sup-SimCSE~\citep{gao-etal-2021-simcse} which leverages annotated natural language inference datasets by using ``entailment'' pairs as positives and ``contradiction'' pairs as hard negatives in the contrastive learning objective. 
As shown in~\reftab{table:embedding}, ACCENT can benefit from the improvement of the sentence embedding method, \ie, using Sup-SimCSE~\citep{gao-etal-2021-simcse}. We support both Sentence-BERT and Sup-SimCSE in our released ACCENT codebase.

\subsection{Error Analysis}
Since ACCENT is a pipeline framework, there is likely error propagation. In section~\ref{sec:ablation}, we rule out the errors introduced by the event-relation extraction component by using human-extracted gold tuples. Results show that ACCENT with gold tuples (see ``Gold Tuples'' in~\reftab{table:ablation}) gives higher correlation with human judgment than ``ACCENT (whole)'' which uses the model-extracted tuples, indicating that ACCENT can benefit from high quality symbolic intermediate representation. We further include a qualitative analysis of the automatically extracted tuples in Appendix~\ref{appendix:examples}, and believe improving the joint event-relation extraction is a worthwhile direction for future work.




\section{Related Work}
\label{sec:related_work}
\noindent
\textbf{Automatic Open-Domain Dialogue Evaluation}\quad
The evaluation of open-domain dialogue systems has long been a challenge due to the system's open-ended goal~\citep{huang2020challenges}, and simply scoring the overall quality is far from enough~\citep{finch-choi-2020-towards}.
Thus, researchers have decomposed the evaluation of open-domain dialogues into multiple facets and developed corresponding automatic evaluation metrics~\citep{pang-etal-2020-towards,mehri-eskenazi-2020-unsupervised}. While aspects like context coherence~\citep{tao2018ruber,ghazarian-etal-2022-deam}, diversity~\citep{hashimoto-etal-2019-unifying}, engagement~\citep{ghazarian2020predictive}, have been systematically studied in the literature, the aspect of commonsense has long been neglected. 

The closest related work is~\citet{zhou-etal-2021-commonsense} which is mainly about commonsense-focused dialogues collection but also 
proposes an automatic metric for commonsense evaluation by training an MLP regressor on both symbolic and neural features. The symbolic features include the numbers of one-hop and two-hop triplets in ConceptNet~\citep{speer2017conceptnet} that can be found between the target response and its dialogue history. 
Although this metric utilizes the CSKB explicitly, it is limited to the direct search with surface form and only considers the number of triplets, and the CSKB used in the work is more about concepts but not event commonsense.

\noindent
\textbf{Joint Event-Relation Extraction}\quad
While event extraction~\citep{ahn-2006-stages}
and relation identification~\citep{do-etal-2011-minimally}
are well-studied, how to jointly acquire them remains a challenge.  
We argue that joint event-relation extraction is an important problem because in practical use cases, 
the input is usually free-form text without pre-extracted events.  
\citet{ZhangLPSL20} is a pioneer work trying to jointly extract event and relation through a pipeline to automatically construct large knowledge graphs. Researchers in this work resort to rule-based methods for event extraction and train a classifier to predict the relation between a pair of events. 

\noindent
\textbf{CSKB Compatibility}\quad
CSKB population enlarges CSKB automatically by adding new links or nodes which are compatible with the commonsense knowledge to the existing CSKB. In~\citet{fang-etal-2021-benchmarking,fang2021discos}, researchers try to add events from a large event knowledge graph to a CSKB. Compatibility test component of ACCENT is relevant to CSKB population task and it is defined in a more general setting where the head and tail of the tuple can be arbitrary events.

\section{Conclusion}
We present ACCENT, an automatic evaluation metric for event commonsense evaluation of open-domain dialogue systems. We show that by using event-relation tuples as the symbolic intermediate representations, ACCENT can effectively utilize the CSKB and achieve a decent correlation with human judgments for dialogue commonsense evaluation. 

\section{Limitations}

In this work, we conduct research on event commonsense of open-domain dialogue systems for the first time. While achieving higher correlations with human judgments than existing baselines, ACCENT has some limitations:

First, the ACCENT framework is based on a fixed set of event relations and the commonsense knowledge in $\text{ATOMIC}_{20}^{20}$ which may fail to cover some potential event commonsense aspects. We believe augmenting the current framework with more commonsense resources is a worthwhile direction for the further improvement of ACCENT.

Second, the event-relation extractor in ACCENT framework is a T5 model fine-tuned in a low resource setting. Although the current model can yield fairly strong performance, it is an important research direction to improve the joint event-relation extraction component because the extracted tuples serve as the symbolic representation for commonsense reasoning in ACCENT framework. Since human extracted tuples are very costly to collect, we hope to explore whether we can improve this component through high-quality synthetic data construction or transfer learning in the future.

\section{Acknowledgments}
We thank the PlusLab members and the anonymous reviewers for their constructive feedback.
This work is supported in part by the DARPA Machine Common Sense (MCS) program under Cooperative Agreement N66001-19-2-4032, and a Meta Sponsored research award. The research is also supported in part by an Amazon Alexa AI gift award.

\bibliography{anthology,custom}


\clearpage
\appendix
\label{sec:appendix}


\section{Commonsense Knowledge Bases}
\label{appendix:cskb}

To endow machines with commonsense reasoning abilities, a growing number of CSKBs are developed through human annotation and information extraction. From earlier CSKBs, ConceptNet~\citep{liu2004conceptnet, speer2017conceptnet} focuses more on taxonomic (\eg, ``IsA'') and lexical (\eg, ``Synonym'', ``RelatedTo'') knowledge; TransOMCS~\citep{ijcai2020p0554} automates the knowledge base construction by leveraging the same limited set of relations defined in ConceptNet.

Recent CSKBs give more focus on event commonsense knowledge. In this work, we select $\text{ATOMIC}_{20}^{20}$~\citep{Hwang2021COMETATOMIC2O} as the knowledge source of ACCENT framework because it is a comprehensive CSKB with rich knowledge regarding how events and human actions are associated with each other. For comparison, ATOMIC~\citep{sap2019atomic}, as the pioneer of $\text{ATOMIC}_{20}^{20}$, consists of only nine relations and therefore poses limitations. Another recent event-centric CSKB is GLUCOSE~\citep{mostafazadeh-etal-2020-glucose}, which however focuses on a specific part of event commonsense (mostly on causal inference) and is less comprehensive and suitable for our work.

\section{Pseudo Code of ACCENT}
\label{appendix:code}
\begin{algorithm}
    \SetKwFunction{genEvents}{generate}
    \SetKwFunction{queryTails}{query}
    \SetKwFunction{checkFormat}{check\_format}
    \SetKwFunction{isEmpty}{is\_empty}
    \SetKwFunction{parse}{parse}
    \SetKwFunction{contradiction}{contradiction}
    \SetKwFunction{max}{max}
    \SetKwFunction{average}{average}
    \SetKwFunction{cosine}{cos}
    \SetKwInOut{KwIn}{Input}
    \SetKwInOut{KwOut}{Output}

    \KwIn{Dialogue history $h$, target utterance $u$, prompt dict $P$, extractor $\mathcal{M}$, dynamic CSKB $\mathcal{C}$, sentence embedder $\mathcal{E}$}
    \KwOut{Event commonsense score $s$}

    tuples $\leftarrow [\ ]$

    \tcp{Event-relation extraction.}
    \ForEach{(\upshape{rel, prompt}) in $\mathcal{P}$}{
        raw\_output $\leftarrow$ $\genEvents(\mathcal{M}, h, u, p)$
        
        \If{$\checkFormat$\upshape{(raw\_output)}}{
            (head, tail) $\leftarrow$ $\parse$(raw\_output)
            
            tuples.append((head, rel, tail))
         }
    }
    
    \tcp{Compatability test.}
    
    \eIf{$\isEmpty$(\upshape{tuples})}{
        \KwRet{0.5}
    }
    {
        tuple\_scores $\leftarrow [\ ]$
        
        \ForEach{\upshape{(head, rel, tail) in tuples}}{
            score $\leftarrow 0$
        
            cs\_tails $\leftarrow$ $\queryTails$($\mathcal{C}$, head, rel)
            
            \ForEach{\upshape{cs\_tail in cs\_tails}}{
                x $\leftarrow$ $\cosine$($\mathcal{E}$(tail), $\mathcal{E}$(cs\_tail))
                
                score $\leftarrow$ $\max$(score, x)
            }
            
            tuple\_scores.append(score)
        }
        
        \KwRet{$\average$\upshape{(tuple\_scores)}}
        
    }

    \caption{ACCENT framework.}
    \label{code:pipeline}
\end{algorithm}

In~\cref{sec:method}, we introduce the symbolic intermediate representation and the two components in ACCENT. Algorithm~\ref{code:pipeline} displays the skeleton of the whole framework.

Line 3-9 in Algorithm~\ref{code:pipeline} show the joint event-relation extraction in ACCENT. We query the event-relation extraction model $\mathcal{M}$ with prompts for each relation. The head and tail events can be parsed from the generated result if it is not ``None'' and follows the pre-defined format, \ie, ``event1: \{head\}; event2: \{tail\}''. Line 16-21 in Algorithm~\ref{code:pipeline} show the compatibility test in ACCENT. Each tuple is given a score based on the maximum cosine similarity between its tail and 
the commonsense tails obtained from the dynamic CSKB $\mathcal{C}$. After calculating scores for each extracted tuple, we average them to get the event commonsense score for the target utterance (Line 24 in Algorithm~\ref{code:pipeline}).

\section{Event Relations}
\label{appendix:relation}
As introduced in~\cref{sec:preliminaries}, ACCENT selects relations from $\text{ATOMIC}_{20}^{20}$ which are related to event commonsense. These event relations can help cover different types of insensibility for event commonsense evaluation of open-domain dialogues. 
\reftab{table:relations} shows examples in DECO where the system response violates event commonsense in terms of different types of event relations. 


Note that although ``Cause'' and ``xReason'' in $\text{ATOMIC}_{20}^{20}$ are also related to event commonsense, we exclude them from the selected subset $\mathcal{\tilde{R}}$. This is because the cause-effect relation can be covered by ``xEffect''/``oEffect'' and tuples with ``Cause'' and ``xReason'' relations take up less than 0.1\% percent of $\text{ATOMIC}_{20}^{20}$. Moreover, we exclude ``IsBefore'' because a tuple with ``IsBefore'' relation can be equivalently converted to a tuple with ``IsAfter'' relation by switching the head and tail. As shown in~\reftab{table:prompt}, for each relation in $\mathcal{\tilde{R}}$, a prompt is manually designed to explain its semantic meaning. These designed prompts give more hints to the pre-trained model and allow a single model to extract tuples for different relations.

\begin{table*}
\resizebox{\textwidth}{!}{%
\centering
\begin{tabular}{clll} 
\toprule
Relation &  & \textbf{Negative Example}                                                                                                                                                                                                  & \textbf{Event-Relation Tuple}                \\ 
\midrule
\makecell{xIntent\\\texttt{Motivation}}        &\makecell{~\\\faIcon{robot}\\~} & \makecell[l]{A: Stay around for a while. The party is warming up.\\B: We'll need to \colorbox{mygreen}{get you some ice cream}, you know,\\\quad~ \colorbox{myyellow}{to warm up your body}.}                   & \makecell[l]{(PersonX gets PersonY some ice cream, xIntent,\\PersonX warms up PersonY's body)}                          \\
\hline
\makecell{xNeed\\\texttt{Need}}        &\makecell{~\\~\\\faIcon{robot}\\~} & \makecell[l]{A: I boated around the world with my husband when\\\quad~  we were younger.\\B: \colorbox{mygreen}{I love boating. I also like to paint}, I just \colorbox{myyellow}{need an}\\\quad~\colorbox{myyellow}{ oxygen mask}. I need a life.}                   & \makecell[l]{(PersonX loves boating, xNeed, PersonX needs\\an oxygen mask), (PersonX likes to paint, xNeed,\\PersonX needs an oxygen mask)}                             \\
\hline
\makecell{xReact, oReact\\\texttt{Reaction}}       &\makecell{~\\~\\~\\~\\\faIcon{robot}\\~} & \makecell[l]{A: That is funny! At work they make me wear a uniform,\\\quad~ boohoo!\\B: That is unfortunate, \colorbox{mygreen}{I actually lost my arm from a car}\\\quad~\colorbox{mygreen}{ accident} so I do not have to.\\A: \colorbox{myyellow}{That is interesting}! Do you know Apple has anything\\\quad~ to say on that?}                   & \makecell[l]{(PersonX loses PersonX's arm from a car accident,\\oReact, PersonY feels interesting)}                        \\
\hline
\makecell{xWant, oWant\\\texttt{Want}}        &\makecell{~\\~\\~\\~\\\faIcon{robot}\\~} & \makecell[l]{A: We don't give bonus every month, but we offer\\\quad~ a semi-annual bonus. And you will receive two\\\quad~ weeks paid vacation a year, as well. Does it suit you?\\B: Yes, thank you. \colorbox{mygreen}{May I ask for an apartment}?\\A: No... \colorbox{myyellow}{I want to take your word on that one}! It'll be \\\quad~ all I need :)}                   & \makecell[l]{(PersonX asks for an apartment, oWant, PersonY\\wants to take PersonX's word)}                         \\
\hline
\makecell{xAttr\\\texttt{Description}}        &\makecell{~\\~\\\faIcon{robot}} & \makecell[l]{A: Are you a vegetarian? I am.\\B: Yes I am. I do not like meat.\\A: \colorbox{myyellow}{I'm a vegetarian} and \colorbox{mygreen}{I love meat}.}                   & (PersonX loves meat, xAttr, PersonX be a vegetarian)                        \\
\hline
\makecell{xEffect, oEffect\\\texttt{Cause-Effect}} &\makecell{~\\~\\\faIcon{robot}\\~}& \makecell[l]{A: How you celebrate your Valentine's Day with\\\quad~ your wife?\\B: I am not sure about you, but \colorbox{mygreen}{my wife is not into}\\\quad~ \colorbox{mygreen}{Valentine's day}... So \colorbox{myyellow}{we celebrate a lot}.}                  & \makecell[l]{(PersonX be not into Valentine's day, xEffect,\\PersonX celebrates a lot)}  \\
\hline
\makecell{HinderedBy\\\texttt{Constraint}} &\makecell{~\\\faIcon{robot}\\~}& \makecell[l]{A: My mom does not bake, she does not even cook.\\B: My mom used to cook for my family, but I think \\\quad~\colorbox{myyellow}{my mom's  got too big} to \colorbox{mygreen}{cook} anything for anymore.}                  & \makecell[l]{(PersonX cooks for family, HinderedBy, PersonX\\gets too big)}  \\
\hline
\makecell{IsAfter\\\texttt{Temporal}}    &\makecell{~\\~\\ \faIcon{robot}\\~} & \makecell[l]{A: Marco has fallen off a ladder. I think he's hurt his\\\quad~ back. What shall we do?\\B: \colorbox{myyellow}{Marco is still on the ladder}, it \colorbox{mygreen}{just got knocked over}.\\\quad~ Marco will not get any sleep.} & \makecell[l]{(PersonX be on the ladder, isAfter, PersonX\\gets knocked over)}                 \\
\hline
\makecell{HasSubEvent\\\texttt{Parent-Child}} &\makecell{~\\~\\~\\\faIcon{robot}\\~}& \makecell[l]{A: Yeah he was an internal medicine practitioner before\\\quad~  he turned to comedy so \colorbox{mygreen}{he attended to the woman}\\\quad~  until medics arrived.\\B: Ohhh I see. I thought he was in the audience when \\\quad~ \colorbox{myyellow}{he was having the seizure}.}                  & \makecell[l]{(PersonX attends to the woman, HasSubEvent,\\PersonX has the seizure)}  \\
\bottomrule
\end{tabular}
}
\caption{Event relations with corresponding negative examples in DECO. \faIcon{robot} denotes responses generated by open-domain dialogue systems. Each example contains events (highlighted with \colorbox{mygreen}{green} and \colorbox{myyellow}{yellow}) which violate event commonsense in terms of the corresponding event relation. Such event commonsense errors can be captured by nonsensical event-relation tuples.} 
\label{table:relations}
\end{table*}

\begin{table*}
\resizebox{\textwidth}{!}{%
\centering
\begin{tabular}{llll} 
\toprule
\textbf{Relation} & \textbf{Semantic Meaning} & \textbf{Designed Prompt} (Extract event1 and event2 from the text where ...)                                                                               \\ 
\midrule
xIntent           & because PersonX wanted          &  event2 shows PersonX's intent for event1.              \\ 
xNeed             & but before, PersonX needed          &  event2 needs to be true for event1 to take place.      \\ 
xReact            & as a result, PersonX feels          &  event2 shows how PersonX reacts to event1.             \\
oReact              & as a result, Y or others feels        &  event2 shows how PersonY reacts to event1.             \\ 
xWant           & as a result, PersonX wants            &  event2 shows what PersonX wants after event1 happens.  \\
oWant           & as a result, Y or others wants            &  event2 shows what PersonY wants after event1 happens.  \\ 
xAttr           & X is seen as            &  event2 shows how PersonX is viewed as after event1.    \\ 
xEffect         & as a result, PersonX will            &  event2 shows the effect of event1 on PersonX.          \\
oEffect         & as a result, Y or others will            &  event2 shows the effect of event1 on PersonY.          \\ 
HinderedBy      & can be hindered by            &  event1 fails to happen because event2.                 \\ 
IsAfter         & happens after            &  event1 happens after event2.                           \\ 
HasSubEvent     & includes the event/action            &  event1 includes event2.                                  \\
\bottomrule
\end{tabular}
}
\caption{Semantic meanings and designed prompts for the selected $\text{ATOMIC}_{20}^{20}$ relations. The semantic meanings are from~\citet{Hwang2021COMETATOMIC2O}.}
\label{table:prompt}
\end{table*}

\section{Additional Details of Data Collection}

\label{appendix:data}
\subsection{Quality Control}
To ensure the annotators have a good understanding of event and event commonsense, we restrict the annotators from English-speaking countries, and those who have finished at least 5,000 HITs with an acceptance rate $>97\%$. The compensation rate for annotators is calculated using a per hour wage of \$16. \footnote{We pay \$1 per HIT for the scoring task and \$3 per HIT for the tuple extraction task. An additional bonus is sent to annotators who successfully pass the training round.}

\begin{figure*}[t]
    \centering 
    \resizebox{1\textwidth}{!}{%
    \includegraphics{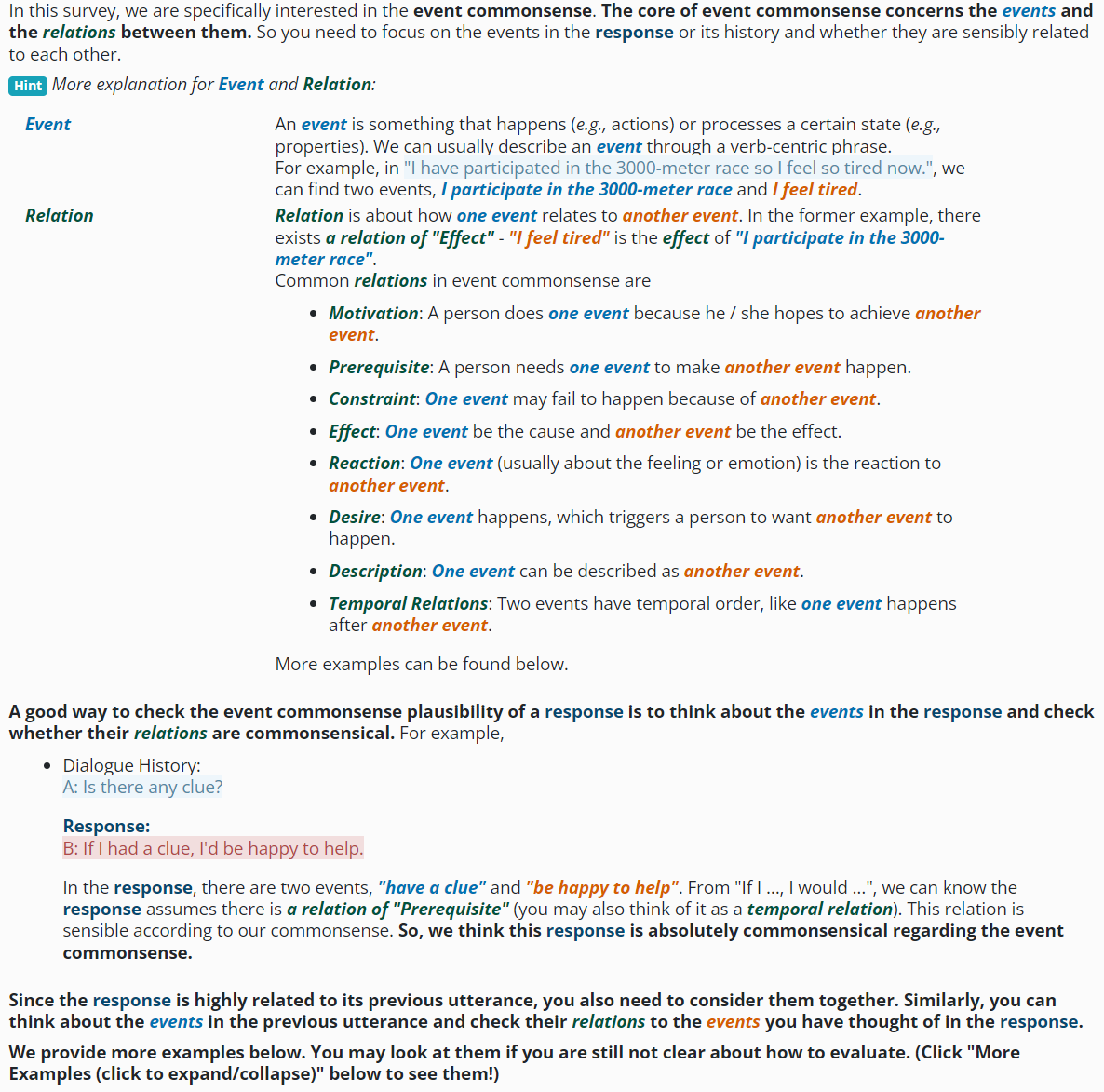}
    }
    \caption{
    Annotation guideline for event commonsense scoring.}
    \label{Fig.scoring_guideline}
\end{figure*}

For commonsense scoring (see~\cref{sec:scoring_collection}), we requested 3 annotators to score each sample, and we instructed them to specifically consider events and their relations in the dialogue to give the event commonsense score. \reffig{Fig.scoring_guideline} shows the annotation guideline we used for event commonsense scoring. We also set a sample for attention check in each HIT. HITs that failed the check were reassigned to other annotators. 

\begin{figure*}[t]
    \centering 
    \resizebox{0.95\textwidth}{!}{%
    \includegraphics{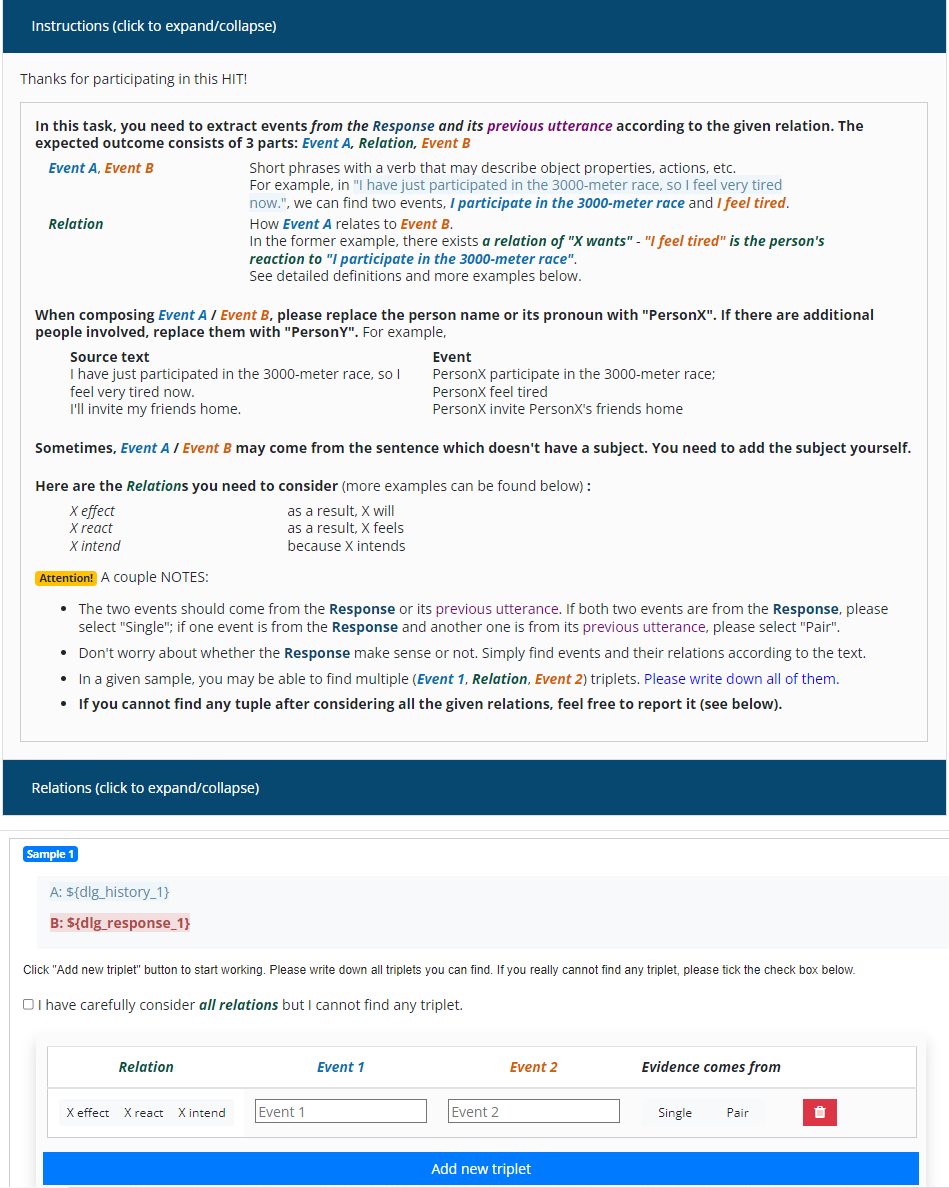}
    }
    \caption{
    Mechanical Turk template used to collect event-relation tuples. This template considers ``xEffect'', ``xReact'', ``xIntent'' relations.}
    
    \label{Fig.amt}
\end{figure*}

For tuple extraction (see~\cref{sec:tuple_collection}), we conducted a training round before the large scale annotation and 8 annotators proceeded to the final round. Each HIT in this task was assigned to 2 individual annotators. The template used to collect event-relation tuples is shown in~\reffig{Fig.amt}. When validating the extracted tuples, 3 annotators judged each tuple, and we achieved Fleiss' Kappa~\citep{fleiss1971measuring} $\kappa = 0.491$ (moderate aggreement). Tuples marked as invalid by the majority vote are not included in the final dataset.

\subsection{Dataset Statistics}
\reftab{table:statistics1} gives the statistics of DECO and ConTurE Subset. Although machine generated responses in DECO are given by advanced open-domain dialogue systems, some event commonsense errors still exist. For ConTurE Subset, we use it to test the generalization ability of different metrics. \reftab{table:statistics2} gives the numbers of human extracted event-relation tuples. Note that the test set of DECO is thoroughly annotated (we consider every relation on each sample) to provide a test set for the joint event-relation extraction task. All the data we collected are in English.

\subsection{License}
The development of DECO and ConTurE Subset is based on the dialogues coming from DailyDialog, PersonaChat, TopicalChat, and ConTurE. PersonaChat\footnote{\url{https://github.com/facebookresearch/ParlAI/blob/main/LICENSE}}, TopicalChat\footnote{\url{https://github.com/alexa/Topical-Chat/blob/master/DATALICENSE}} and ConTurE\footnote{\url{https://github.com/alexa/conture/blob/main/DATALICENSE}} are licensed. We ensure that we did not violate any license conditions when developing our datasets.

\begin{table}
\centering
\resizebox{\columnwidth}{!}{%
\begin{tabular}{llll} 
\toprule
\textbf{Dataset} & \textbf{Size} & \begin{tabular}[c]{@{}l@{}}\textbf{Average~}\\\textbf{score}\end{tabular} & \begin{tabular}[c]{@{}l@{}}\textbf{Average \# tokens }\\\textbf{in target response}\end{tabular}  \\ 
\midrule
DECO             & 300           & 3.39                                                                      & 17.6                                                                                              \\
ConTurE Subset   & 100           & 2.88                                                                      & 15.7                                                                                              \\
\bottomrule
\end{tabular}
}
\caption{Statistics of collected dialogue in event commonsense evaluation datasets.}
\label{table:statistics1}
\end{table}

\begin{table*}
\centering
\resizebox{\textwidth}{!}{%
\begin{tabular}{lllllllllllll} 
\toprule
                 & xIntent & xNeed & xReact & oReact & xWant & oWant & xAttr & xEffect & oEffect & HinderedBy & IsAfter & HasSubEvent  \\ 
\midrule
Train (few-shot) & 20      & 20    & 24     & 30     & 22    & 22    & 25    & 37      & 22      & 29         & 26      & 30           \\
Test             & 37      & 35    & 21     & 50     & 29    & 20    & 46    & 71      & 32      & 20         & 61      & 45           \\
\bottomrule
\end{tabular}
}
\caption{Statistics of the collected event-relation tuples. We collect around 30 tuples for each relation from DECO training set to train the event-relation extractor in the few-shot learning setting. The test set of DECO is thoroughly annotated to provide a test set for the joint event-relation extraction task.}
\label{table:statistics2}
\end{table*}

\section{Additional Details of Experiment}
\label{appendix:implementation}
This section describes more details about baseline implementation, applying ACCENT to CSKB benchmark, and computational resources. The implementation details of the proposed ACCENT framework are discussed in~\cref{sec:implementation}

\subsection{Baseline Implementation}
\label{appendix:baseline}
We compare ACCENT with 5 baseline metrics on event commonsense evaluation. All the metrics are tested on DECO test set and ConTurE Subset. For metrics which require training, the training set of DECO is used, and we split 20\% data for validation. The implementation details are as follows:

\begin{itemize}
    \item{\textbf{FED-understandable/appropriate}: We use their released model\footnote{\url{https://github.com/Shikib/fed}}.}
    \item{\textbf{Cross-encoder}: 
    We use the cross-encoder with a regression head implemented in the Sentence-Tranformers library~\citep{reimers-gurevych-2019-sentence}. We use BART as the backbone model to align with the COMET model used in our method. For hyperparameter search, we fine-tune the cross-encoder for 10 epochs with the batch size of 8 and the learning rate from \{1e-5, 3e-5, 5e-5\}.}
    \item{\textbf{Cross-encoder (COMET)}: We use the off-the-shelf COMET model trained on $\text{ATOMIC}_{20}^{20}$\footnote{\url{github.com/allenai/comet-atomic-2020}} as the backbone model. Other implementation details are the same with \textbf{Cross-encoder}.}
    \item{\textbf{MLP regressor}: Since the code of~\citet{zhou-etal-2021-commonsense} is not publicly available, we produce the results using our own implementation based on scikit-learn\footnote{\url{https://scikit-learn.org/stable/modules/generated/sklearn.neural_network.MLPRegressor.html}}. Our implementation is available in our released codebase.}
    
\end{itemize}

For event-relation extraction baselines, the implementation details are as follows:
\begin{itemize}
    \item{\textbf{ASER Extractor}: We use their provided code\footnote{\url{https://github.com/HKUST-KnowComp/ASER}} to extract events. The neural classifier for relation prediction is trained on the human annotated tuples in DECO training set.}
    \item{\textbf{CSKB Search}: We search tuples related to the target response and its previous response in $\text{ATOMIC}_{20}^{20}$ following the CSKB search pipeline described in~\citet{zhou-etal-2021-commonsense}. A potential concept set is built from the utterances by identifying nouns, verbs, and adjectives that are not stopwords through part-of-speech (POS) tagging and lemmatizing them. We return tuples whose head and tail both contain words in the concept set as the search result.}
\end{itemize}

For CSKB population baselines, we use the implementation in~\citet{fang-etal-2021-benchmarking}\footnote{\url{https://github.com/HKUST-KnowComp/CSKB-Population}}. For the backbone model, we use $\text{RoBERTa}_\text{LARGE}$ which has roughly the same parameter budget with COMET. We train all models for 1 epoch with the batch size of 64. The learning rate is searched from \{1e-7, 1e-5, 1e-3\} on the validation set.

\subsection{Applying ACCENT to CSKB Benchmark}
\label{sec:population_detail}
In~\cref{sec:experiment} Setup 3, we apply the compatibility test approach in ACCENT to a CSKB benchmark. Such an application is seamless because the compatibility test also assigns a score to each tuple, and tuples which receive higher compatibility scores are naturally more suitable to populate the CSKB. We train the COMET model on the positive samples in the training set of the benchmark dataset for 1 epoch with the batch size of 64. The learning rate is searched from \{1e-7, 1e-5, 1e-3\} on the validation set. Note that our approach does not require any negative sample in the training stage. It also does not need the event graph information provided in the benchmark dataset, but results in \reftab{table:population} shows that our method outperforms baselines which require manually created negative data and take in graph information.

\subsection{Computational Resources}
We run \textsc{BertSAGE} and \textsc{KG-BertSAGE} for the CSKB benchmark experiments on a single Nvidia V100 GPU with 32 GB memory where these models require large memory consumption in the run time. The rest of experiments is done on a single Nvidia A10 GPU with 24 GB memory.

Note that although we develop the ACCENT framework based on large language models, the only part which requires training is the T5 model (with 223M parameters) for event-relation extraction. As discussed in~\cref{sec:eventualization}, the model is fine-tuned in a low resource setting and the training takes less than 0.5 GPU hour.

\section{More Discussion of the Compatibility Test Approach in ACCENT}
\label{appendix:compatibility}
\begin{table}
\centering
\begin{tabular}{lcc} 
\toprule
                & \textbf{Pearson} & \textbf{Spearman}  \\ 
\midrule
COMET (neural)  & 0.14             & 0.25               \\
ACCENT approach & \textbf{0.40}             & \textbf{0.42}               \\
\bottomrule
\end{tabular}
\caption{Correlations between human judgments and different compatibility test approaches with human-extracted tuples on DECO test set.}
\label{table:compatibility}
\end{table}

\begin{table*}
\resizebox{\textwidth}{!}{%
\centering
\begin{tabular}{lllll} 
\toprule
\textbf{Dialogue}                                                                                                                                                                                       & \textbf{Tuple}                                                                         & \textbf{Human} & \textbf{\makecell[l]{COMET \\(nueral)}} & \textbf{\makecell[l]{ACCENT \\approach}}  \\ 
\midrule
\multirow{3}{*}{\makecell[l]{A: I work in the bakery and eat all my favorite\\ ~~~~~cupcakes. What do you do?\\ \textbf{B: I actually just got fired for a mistake I made.}}} & (PersonX makes a mistake, xEffect, PersonX gets fired)                                 & /              & 0.33                    & 0.63                      \\
                   & (PersonX gets fired, isAfter, PersonX makes mistake)                                   & /              & 0.12                    & 0.68                      \\
                   & (PersonX gets fired, HasSubEvent, PersonX makes mistake)                               & /              & 0.18                    & 0.66                      \\ 
\cline{2-5}
                   & \textbf{Average}                                                                       & 0.80           & 0.21                    & 0.66                      \\ 
\midrule
\multirow{4}{*}{\makecell[l]{A: Yeah winter is coming soon. It gonna be\\ ~~~~~really cold.\\ \textbf{B: I know I know. I want to live in a cold place}\\ \textbf{~~~~~before I go full on winter.}}} & \makecell[l]{(PersonX wants to live in a cold place, xIntent, PersonX intends to \\go full on winter)} & /              & 0.06                    & 0.59                      \\
                   & (PersonX goes full on winter, xNeed, PersonX lives in cold place)                      & /              & 0.53                    & 0.39                      \\
                   & (PersonX goes full on winter, isAfter, PersonX lives in a cold place)                  & /              & 0.95                    & 0.55                      \\
                   & \makecell[l]{(PersonX knows winter is coming, HasSubEvent, PersonX wants to\\ live in a cold place)}   & /              & 1.63                    & 0.69                      \\ 
\cline{2-5}
                   & \textbf{Average}                                                                       & 0.40           & 0.79                    & 0.56                      \\
\bottomrule
\end{tabular}
}
\caption{Examples of results given by different compatibility test approaches. Only $u_{n-1}$ in the dialogue history is shown. Human annotations are normalized to the range of (0, 1) and scores given by ``COMET (neural)'' are scaled up by 100 times for better demonstration. Compatibility scores for each tuple are averaged to get the final score for the target response (in \textbf{bold} font).}
\label{table:compatibility2}
\end{table*}

ACCENT checks whether an event-relation tuple $(h,r,t)$ is compatible with the commonsense knowledge by comparing the similarity between $t$ and commonsense tails generated by the Dynamic CSKB (COMET). Ablation results in~\reftab{table:ablation} show that the compatibility test approach in ACCENT yields better performance than the ``COMET (neural)'' alternative which also uses the COMET model. To exclude the potential noise introduced by the automatically extracted tuples, we further compare these two methods using human-extracted tuples on DECO test set. Results in~\reftab{table:compatibility} demonstrate that the conclusion still holds under this experimental setting. \reftab{table:compatibility2} gives two samples with a breakdown of tuple results. Compared with the compatibility scores given by ACCENT approach, the scores given by ``COMET (neural)'' are less comparable among tuples with different relations, thus making this method unsuitable for ACCENT framework.

\section{Error Analysis}
\label{appendix:examples}
We conduct a qualitative analysis of the event-relation component in ACCENT. \reftab{table:examples} shows some examples of the extracted tuples. While most of the head and tail events are nicely composed and capture the major information in the given text, they are not perfect. Multiple participants involved in the dialogue further increase the difficulty of the task. We note that the model sometimes confuses the multiple participants in the dialogue and makes mistakes when using ``PersonX'' or ``PersonY''. For example, in the third sample of \reftab{table:examples}, the model confuses different participants since the subjects of ``raise money'' and ``go tomorrow for treatment'' should be different. Such confusion will lead to the wrong tuples which cannot truly reflect the meaning of the dialogue. Also, identifying relation from the given dialogue is challenging. Although we include negative samples (dialogues that do not have a certain relation) when fine-tuning T5, errors still exist (\eg, the tuple with ``xAttr'' relation in the third sample of \reftab{table:examples}).

\begin{table*}
\centering
\resizebox{\textwidth}{!}{%
\begin{tabular}{l} 
\toprule
\textbf{Automatically extracted event-relation tuples}                                                                                                                                                                                                                                                                                                           \\ 
\midrule
\begin{tabular}[c]{@{}l@{}}Dialogue History:\\A: Cool. Are you religious? I have not told my parents I am wiccan yet.\\B: I am the wife of a pastor. Spirituality is important in our home.\\A: Oh. I grew up in a un religious home.\\B: God is always around, I hope he does not mind that I do not like cleaning.\\Target Response:\\A:~I'm not religious. I'm a Christian.\end{tabular}  \\
\color{blue}{\makecell[l]{(PersonX not be religious, xAttr, PersonX be a Christian), (PersonX grew up in un religious home, xEffect,\\ PersonX be a Christian), (PersonX like cleaning, isAfter, PersonX be a Christian), (PersonX like cleaning,\\ HasSubEvent, PersonX be a Christian), (PersonX like cleaning, HinderedBy, PersonX be a Christian)}}                                                                     \\ 
\hline
\begin{tabular}[c]{@{}l@{}}Dialogue History:\\A: I'm dreadfully sorry for you. \\B: Thank you very much for your kindness.\\A: Is there any clue?\\Target Response:\\B:~If I had a clue, I'd be happy to help.\end{tabular}                                                                                                                                                                 \\
\color{blue}{\makecell[l]{(PersonX have a clue, xAttr, PersonX be happy to help), (PersonX has a clue, xReact, PersonX be happy to help),\\ (PersonX has a clue, xEffect, PersonX be happy to help), (PersonX asks PersonY for a clue, oEffect, PersonY \\is happy to help), (PersonX help PersonY, HinderedBy, PersonX have a clue)}}                                                                                       \\
\hline
\begin{tabular}[c]{@{}l@{}}Dialogue History:\\A: you mean if they catch the murderer. I'm scared stiff about going into the city center at night.\\B: There must have been some good news in the newspaper. I can't believe that none of the news stories make\\ you happy or excited.\\A: Well, there was one good piece of news. You remember the local girl who was dying of a rare blood disease?\\B: Yes. Her parents were raising money to have her treated in the united states.\\Target Response:\\A: Well, they've got the money and she's going tomorrow for treatment.\end{tabular}                                                                                                                                                                 \\
\color{blue}{\makecell[l]{(PersonX raise money to have PersonY treated in United States, xAttr, PersonX go tomorrow for treatment),\\(PersonX raises money to have PersonY treated in United States, xReact, PersonX feels good), (PersonX raise \\money to have PersonY treated in United States, xEffect, PersonX go tomorrow for treatment), (PersonX raise \\money to have PersonY treated in United States, xWant, personX go tomorrow for treatment), (PersonX raise \\money to have PersonY treated in United States, xIntent, PersonX go tomorrow for treatment), (PersonX raise \\money to have PersonY treated in United States, oEffect, PersonY go tomorrow for treatment), (PersonX raises \\money to have PersonY treated in United States, oReact, PersonY feels happy), (PersonX raise money to have \\PersonY treated in United States, oWant, PersonY go tomorrow for treatment), (PersonX go to the United States, \\isAfter, The parents raise money to have the girl treated)}}                                                                                       \\
\hline
\begin{tabular}[c]{@{}l@{}}Dialogue History:\\A: Yeah I know, he did a great job for the limited time, can you believe that 20th Century Fox let George Lucas \\keep the merchandising rights, that's what made him a billionaire, so crazy I don't think they would do that\\ever again.\\B: I think I can believe it. Good for him, what does the corporation need with something they didn't invent \\anyways?\\A: No I agree with you. Are you going to watch the Academy Awards by the way? They give out 24 Awards\\and I think that the first broadcast was back in 1930, which is wild\\B: Yeah, so like the academy awards totally suck. Does anyone actually give a bleep what they think?\\Target Response:\\A: Well I can't speak for the other commenters but I'm sure there are people who give them a bleep can help \\winners go on to win awards and make money.
\end{tabular}                                                                                                                                                                 \\
\color{blue}{\makecell[l]{(PersonX give a bleep, HasSubEvent, PersonX help winners go on to win awards)}}                                                                                       \\
\bottomrule
\end{tabular}
}
\caption{Some examples of the automatically extracted event-relation tuples. These tuples function as the intermediate symbolic representation in ACCENT framework.}
\label{table:examples}
\end{table*}

\end{document}